\documentclass{article} 
\usepackage{iclr2020_conference,times}

\usepackage{amsmath,amsfonts,bm}

\def\eqref#1{equation~\ref{#1}}

\def\1{\bm{1}}

\DeclareMathAlphabet{\mathsfit}{\encodingdefault}{\sfdefault}{m}{sl}
\SetMathAlphabet{\mathsfit}{bold}{\encodingdefault}{\sfdefault}{bx}{n}

\usepackage{hyperref}
\usepackage{url}

\usepackage{graphicx}   
\usepackage{subfigure}
\usepackage{multirow}
\usepackage{diagbox}
\usepackage{color}
\usepackage{bbm}
\usepackage{multicol}
\usepackage{blindtext}
\usepackage{amsmath,amsfonts,amssymb}
\usepackage{bm}
\usepackage{mathtools}
\usepackage{wrapfig}
\usepackage{wrapfig,lipsum,booktabs}
\usepackage{algorithm}
\usepackage{algpseudocode}
\usepackage{url}
\usepackage[normalem]{ulem}

\usepackage{soul}

\DeclareMathOperator*{\transformer}{TransformerEnc}

\DeclareMathOperator*{\pool}{Pool}
\DeclareMathOperator*{\bilinear}{BiLnr}
\DeclareMathOperator*{\docfreq}{DF}
\DeclareMathOperator*{\neighbor}{Nb}
\DeclareMathOperator*{\gpath}{PLen}

\graphicspath{{./figs/}}

\title{Exploiting Structured Knowledge in Text via \\ Graph-Guided Representation Learning}

\author{Tao Shen\thanks{Work done while the author was an intern at Microsoft.}\\
	University of Technology Sydney\\
	\texttt{tao.shen@student.uts.edu.au} \\\And
	Yi Mao, Pengcheng He\\
	Microsoft Dynamics 365 AI\\
	\texttt{\small\{maoyi,penhe\}@microsoft.com} \\\AND
	Guodong Long \\
	University of Technology Sydney \\
	\texttt{guodong.long@uts.edu.au} \\\And
	Adam Trischler \\
	Microsoft Research, Montr\'eal~~~~~~~~~ \\
	\texttt{adtrisch@microsoft.com} \\\And
	Weizhu Chen\\
	Microsoft Dynamics 365 AI\\
	\texttt{\small\{wzchen\}@microsoft.com} \\
}

\renewcommand{\headrulewidth}{0pt}  

\iclrfinalcopy 
\begin{document}

\maketitle

\begin{abstract}
In this work, we aim at equipping pre-trained language models with structured knowledge. We present two self-supervised tasks learning over raw text with the guidance from knowledge graphs.  Building upon entity-level masked language models, our first contribution is an entity masking scheme that exploits relational knowledge underlying the text. This is fulfilled by using a linked knowledge graph to select informative entities and then masking their mentions. In addition we use knowledge graphs to obtain distractors for the masked entities, and propose a novel distractor-suppressed ranking objective which is optimized jointly with masked language model. In contrast to existing paradigms, our approach uses knowledge graphs implicitly, only during pre-training, to inject language models with structured knowledge via learning from raw text. It is more efficient than retrieval-based methods that perform entity linking and integration during finetuning and inference, and generalizes more effectively than the methods that directly learn from concatenated graph triples. Experiments show that our proposed model achieves improved performance on five benchmark datasets, including question answering and knowledge base completion tasks. 
\end{abstract}

\section{Introduction} \label{sec:introduction}

Self-supervised pre-trained language models (LMs) like ELMo \citep{pt/elmo} and BERT \citep{pt/bert} learn powerful contextualized representations. With task-specific modules and finetuning, they have achieved state-of-the-art results on a wide range of natural language processing tasks. 
Nevertheless, open questions remain about what these models have learned and improvements can be made along several directions. 
One such direction is, when downstream task performance depends on structured relational knowledge -- the kind modeled by knowledge graphs\footnote{``Knowledge graph (KG)'' and ``knowledge base (KB)'' are exchangeable in this paper, denoting triple-formatted curated graph.} (KGs) --
directly finetuning a pre-trained LM often yields sub-optimal results, even though some works \citep{uc/petroni2019language,uc/feldman2019commonsense} show pre-trained LMs have been partially equipped with such knowledge. 

\begin{figure}[t]\small
	\centering
	\includegraphics[width=0.6\textwidth]{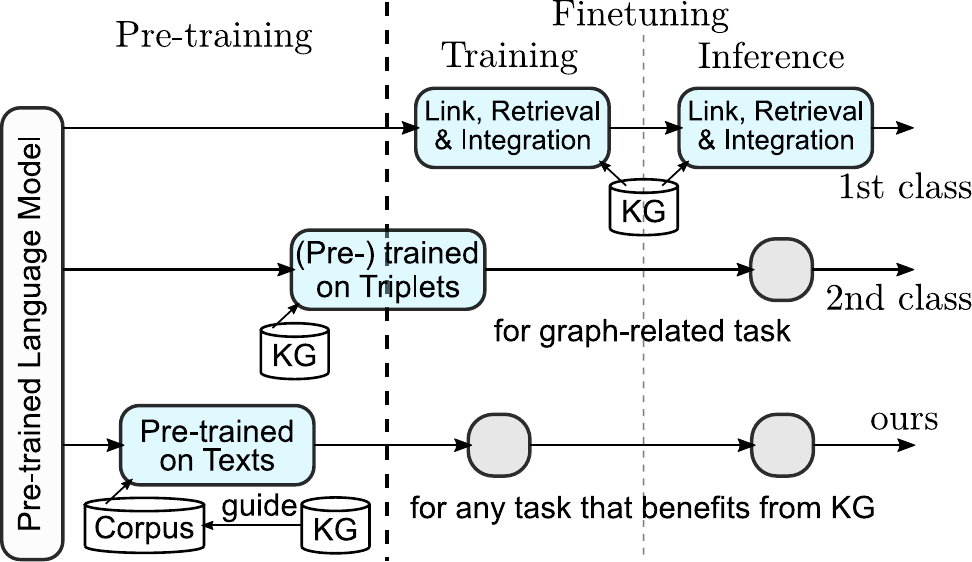}
	\caption{\small Taxonomy of different approaches to integrating pre-trained LMs with knowledge graphs.}
	\label{fig:comp_pipeline} 
	\centering
\end{figure}

To address this shortcoming, several recent works attempt to integrate KGs into pre-trained LMs. These approaches can be coarsely categorized into two classes, as shown in Figure~\ref{fig:comp_pipeline}. 
The first line of methods retrieve a KG subgraph \citep{uc/liu2019kbert,uc/lin2019kagnet,uc/lv2019graph} and/or pre-trained graph embeddings \citep{uc/thu2019ernie,uc/peters2019knowledge} via entity linking during both training and inference on downstream tasks.
While these methods inject domain-specific knowledge directly into language representations, they rely heavily on the performance of the linking algorithm and/or the quality of graph embeddings. Graph embeddings, to be tractable over large-scale KGs, are often learned using shallow models (e.g., TransE \citep{gnn/TransE}) with limited expressive power. 
Besides, the linking and retrieval invoked during both finetuning and inference are costly, hence limit these methods' practicality.

The second class of methods \citep{uc/bosselut2019comet,uc/malaviya2019exploiting,yao2019kgbert} use contextualized representations from pre-trained LMs to enrich graph embeddings and thus alleviate graph sparsity issues. This is especially helpful in the case of commonsense KGs (e.g., ConceptNet \citep{db/conceptnet}) that consist of non-canonicalized text and hence suffer from severe sparsity \citep{uc/malaviya2019exploiting}. 
Specifically, these methods usually feed concatenated triples (e.g., [\textsc{Head}, \textsl{Relation}, \textsc{Tail}]) into LMs for training or finetuning. 
The drawback is that focusing on knowledge base completion tends to over-adapt the models to this specific task, which comes at the cost of generalization. 

In this work, we equip masked language models (MLMs), e.g. BERT, with structured knowledge via self-supervised pre-training on raw text. 
Compared to the first class, we expose LMs to structured information only during pre-training, thus circumvent costly knowledge retrieval and integration in finetuning and inference. Also the dependency on the performance of linking algorithm is greatly reduced. Compared to the second class, we learn from free-form text through MLMs rather than triples, which fosters generalization on other downstream tasks. 

Specifically, given a corpus of raw text and a KG, two KG-guided self-supervision tasks are formulated to inject structured knowledge into MLMs. 
First, taking inspiration from Baidu-ERNIE \citep{uc/baidu2019ernie}, we reformulate the masked language modeling objective to an \emph{entity}-level masking strategy, where entities are identified by linking their text mentions to \emph{concepts/phrases} in a commonsense KG or \emph{named entities} in an ontological KG \citep{db/freebase}. The role of KG here is to provide a ``vocabulary'' of entities to be masked. To further exploit implicit relational and logical information underlying raw text, we design a KG-guided masking scheme that selects informative entities by considering both document frequency and mutual reachability of the entities detected in the text. 
In addition to the new entity-level MLM task above, a novel distractor-suppressed ranking task is proposed. Negative entity samples are derived from the KG and used as distractors for the masked entities to make the learning more effective. 

Note that our approach never observes the KG directly, through triples or other forms. Rather, the KG plays a guiding role in our proposed tasks. Its guidance helps the model exploit the text corpus more effectively, as verified in the experiments. 
If a downstream task can benefit from explicit exposure of KG, a method by \citet{uc/feldman2019commonsense} can be used to transform KG triples into natural grammatical texts that can be passed into our model. 

We evaluate our method on five benchmarks, including question answering and knowledge base completion (KBC) tasks. Results show our method achieves state-of-the-art or competitive performance on all benchmarks, followed by analyses on the effects of various modeling choices.

\begin{figure*}[t]\small
	\centering
	\includegraphics[width=0.95\textwidth]{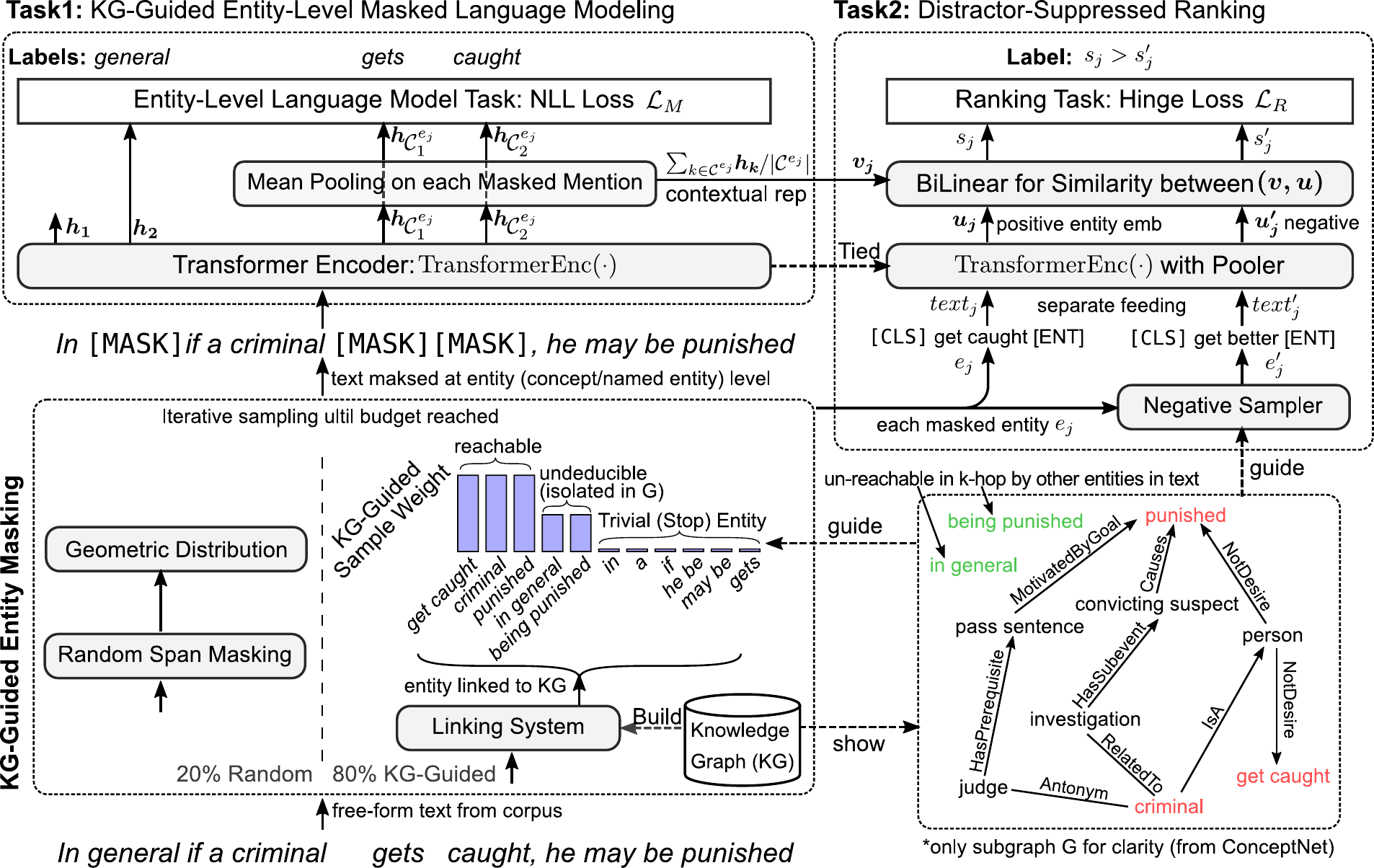}
	\caption{\small \textbf{G}raph-guided Masked \textbf{L}anguage \textbf{M}odel (GLM) with two self-supervised tasks for MLM training.}
	\label{fig:app_illu} 
	\centering
\end{figure*}

\section{Vanilla Masked Language Model}\label{sec:vanilla_mlm}

To ground our approach, this section summarizes MLMs for pre-training bidirectional Transformers \citep{pt/bert}. 
Compared to causal LMs \citep{pt/elmo} trained unidirectionally, MLMs randomly mask some tokens and predict the masked tokens by considering their context on both sides. Formally, given a piece of text $U$, a tokenizer, e.g. BPE \citep{basic/bpe}, is used to produce a sequence of tokens $[w_1, \dots, w_n]$. A certain percentage of the original tokens are then masked and replaced: of those, 80\% with special token \texttt{[MASK]}, 10\% with a token sampled from the vocabulary $\mathbb{V}$, and the remaining kept unchanged. The masked sequence, denoted as $[w_1^{(m)}, \dots, w_n^{(m)}]$, is passed into a Transformer encoder to produce contextual representations for the sequence:
\begin{align}
\bm{H} = \transformer\left(\left[w_1^{(m)}, \dots, w_n^{(m)}\right]\right), \label{eq:trans_enc}
\end{align}
where $\bm{H} \in \mathbb{R}^{d_h\times n}$ and $d_h$ denotes the hidden size. 
The training loss $\mathcal{L}_{M}$ for the MLM task is defined as
\begin{align}
\mathcal{L}_{M} = - \dfrac{1}{|\mathcal{M}|} \sum_{i \in \mathcal{M}} \log P(w_i|\bm{H}_{:,i}), \label{eq:loss_mlm}
\end{align}
where $\mathcal{M}$ denotes the set of masked token indices, and $P(w_i|\bm{H}_{:,i})$ is the probability of predicting the original token $w_i$ given the representations computed from the masked token sequence.

\section{Graph-guided Masked Language Model} 

This section begins with a description of entity-level masked language modeling. 
Section~\ref{sec:node_lm} proposes a KG-guided entity masking scheme for the entity-level MLM task. 
A novel distractor-suppressed ranking task is presented in Section~\ref{sec:lm_ge}, which operates on the masked entities and their negative entity samples from the KG. 
We use a multi-task learning objective combining the two tasks above to jointly train our proposed \textbf{G}raph-guided Masked \textbf{L}anguage \textbf{M}odel (GLM). 
An illustration of the GLM is shown in Figure~\ref{fig:app_illu}.

\subsection{Entity-Level Masked Language Model} \label{sec:data_format}
As aforementioned, directly training an MLM with graph triples learns structured knowledge at the cost of the model's generalization to tasks involving natural text such as question answering.
Inspired by distantly supervised relation extraction \citep{mintz2009distant} which assumes that any sentence containing two entities can be used to express the relation between these two entities in a KG, we argue that it is possible for an MLM to learn structured knowledge from raw text if guided properly by a KG. 

Roughly speaking, we take detected entity mentions as masking candidates, where the entity can be a concept/phrase in a commonsense KG or a named entity in an ontological KG. The intuition is that the mentions in text often represent knowledge-grounded, semantically meaningful text spans.  
Formally, we first use a KG to provide a vocabulary of entities for building an entity linking system. We then detect all entity mentions appearing in a piece of text $U$ from a corpus. This leads to a set of linked entities $\mathcal{E} = \{e_1, e_2, \dots\} \triangleq \{e | e \in \text{KG} \wedge e \in U \}$ with  $\mathcal{C}^{e_j}$ being the corresponding token indices in $U$ for each entity mention $e_j$. 

The idea of entity-level masking is not new. For example, \citet{uc/baidu2019ernie} and \citet{pt/spanbert} randomly mask entity candidates under uniform distribution for training MLMs. We take this idea further by building our masking scheme with the guidance of a KG, as explained below.

\subsection{KG-Guided Entity Masking Scheme} \label{sec:node_lm}

In this section, we develop a new entity masking scheme to facilitate structured knowledge learning for MLMs. It explores implicit relational information underlying raw text by the guidance of a KG, and is shown to mask more informative entities compared to the previous random approach.

In particular, the scheme is designed to avoid or reduce masking two types of entities: trivial and undeducible. 
Trivial entities, such as \textit{have been} and \textit{what do} in ConceptNet, are ubiquitous in corpora, since they are used to compose sentences. However, they express bare semantics and function similarly to stop words. 
On the other hand, an undeducible entity is defined as an entity that is hardly reached from any other entities detected in the same text, within certain hops over the linked KG. Examples include general modifiers and ambiguous entity linking results, as shown with green in Figure~\ref{fig:app_illu}.

Given a masking budget (e.g., 20\% of total tokens in our setting), we sample token spans iteratively as follows until the budget is reached: 1) 20\% of the time we sample a random token span under a geometric distribution with $p\!=\!0.2$, and 2) 80\% of the time we sample an entity mention from the candidates detected in \S\ref{sec:data_format}. The probability to sample an entity mention $\mathcal{C}^{e_j}$ is defined as
\begin{align}
 P(\mathcal{C}^{e_j})  \propto \mathbb{I}_{\{\docfreq(e_j)< \text{R}_{\text{thresh}}\}}  \times  \Big[\big|\neighbor(e_j)\big|\Big]^{\text{R}_{\text{max}}}_{\text{R}_{\text{min}}},  \label{eq:prob_heuristic}
\end{align}
where
$ \neighbor(e) \triangleq  \{e' |~ \gpath(e'\leftrightarrow e) < \text{R}_{\text{hop}} \wedge e' \in \mathcal{E}  \}$. The term $\docfreq(\cdot)$ denotes document frequency, $\gpath(e \leftrightarrow e')$  is the length of the shortest undirected path between the two entities, $|\cdot|$ denotes the set size, and $[x]_{a}^{b} \triangleq \max(a,\min(x,b))$.

Note, the first part in Eq.(\ref{eq:prob_heuristic}) is designed to eliminate trivial entities that frequently appear. The second part measures whether an entity can be reached from other entities detected in the same text within $\text{R}_{\text{hop}}$-hops, and assigns a higher sampling weight to an entity (e.g., \textit{criminal} in Figure \ref{fig:app_illu}) that could more easily be inferred by others.  By guiding the model to favor masking deducible but non-trivial entities, this scheme facilitates the MLM ingesting relational knowledge into representation learning. 
$\text{R}_{\text{hop/thresh/min/max}}$ are hyperparameters that trade off between trivial and undeducible entities.  

Finally, it is worth noting, frequently appearing entities that are excluded via $\mathbb{I}_{\{\cdot\}}$ in Eq.(\ref{eq:prob_heuristic}) can still be masked via 20\% random span masking budget, but now with much smaller probabilities.

\subsection{Distractor-Suppressed Ranking Task} \label{sec:lm_ge}

Empowered by the informative entity-level masks, it is natural to extend the MLM with ``negative'' entities sampled from the KG, by treating the masked entities as ``positive''. 
It has been shown that negative sampling is especially useful for structured knowledge learning in graph embedding approaches \citep{gnn/RotatE,cai2018kbgan}, but how to effectively integrate negative samples from KGs into MLMs remains open. 

Recently, \citet{ye2019align} propose to mask one entity mention in a sentence, and then formulate a multiple-choice QA task for structured knowledge learning, by treating the masked sentence as the question, and the masked entity plus its negative samples as answer candidates. However, this model 
does not quite match the MLM since only one entity can be masked in a text.

Here we propose a distractor-suppressed ranking objective that operates on each pair of a masked entity from \S\ref{sec:node_lm} and its negative sample from the KG. The negative sample can be viewed as a distractor. 
We use a Transformer encoder to separately produce the embeddings of positive and negative entities using their associated node contents in the KG. We then contrast the positive and negative \emph{entity embeddings}, $\bm{u}$ and $\bm{u}'$, against the \emph{masked entity mention's contextual representation}, $\bm{v}$, using vector similarity as plausible scores for the both entities. 

Specifically, given a set of masked entities from \S\ref{sec:node_lm}, $\mathcal{E}^p = \{e_1, \dots, e_m\} \subseteq \mathcal{E}$, with the corresponding entity mentions $\mathcal{C}^{e_j}$, we gather the contextual representation for each masked entity mention, by mean-pooling over representations of its composite tokens, where the representations are generated by the Transformer encoder of the MLM:
\begin{align}
\bm{v_j} = \dfrac{1}{|\mathcal{C}^{e_j}|} \sum_{k \in \mathcal{C}^{e_j}} \bm{H}_{:,k},~ j = 1, \dots
\end{align}
Here $\bm{v_j}$ is the resulting contextual representation for $e_j$.  Since each entity's original mention is invisible to the encoder, $\bm{v_j}$ is rich in contextual features. 

We then sample negative sample(s) from the KG for each $ e_j \in \mathcal{E}^p$ and derive a set of positive-negative entity pairs $\{(e_j, e'_j)\}_{j=1}^{m}$. 
In particular, given a positive entity $e_j$, the sampling method randomly selects an entity $e'_j$ from the KG as a negative sample. The sampling favors its sibling entities with the same relation, whose sample weights are twice than the others. 
This is similar to \citet{ye2019align} and aims to provide strong distractors. 
Then, another Transformer encoder separately encodes positive and negative entities, which is parameter-tied with the MLM in \ref{sec:node_lm} but uses distinct position embeddings. To distinguish entity text coming from KG's node or natural text, we append a special token to the entity text, i.e., $text_j$ = \texttt{[CLS]} + $e_j$ + \texttt{[ENT]}. We pass $text_j$ into the encoder to obtain the entity embedding for $e_j$, i.e.,
\begin{align}
\bm{u_j} = \pool(\transformer(text_j)). \label{eq:lm_aug_ge}
\end{align}
Here, $\pool(\cdot)$ denotes collecting the contextualized embedding from the \texttt{[CLS]} token, as in \citet{pt/bert}. The resulting $\bm{u_j} \in \mathbb{R}^{d_h}$ is an LM-augmented entity embedding for $e_j$. We apply the same process to $e'_j$ to obtain negative entity embedding $\bm{u'_j}$. 

The procedure above yields a set of tuples, $\{(\bm{v_j}, \bm{u_j}, \bm{u'_j})\}_{j=1}^m$. 
Finally, a BiLinear layer (abbrv. $\bilinear$) is used as a parameterized metric to calculate a similarity score between $\bm{v_j}$ and $ \bm{u_j}$ (or $ \bm{u'_j}$). The score is
\begin{align}
&s_j = \bilinear(\bm{v_j}, \bm{u_j}),~~s'_j = \bilinear(\bm{v_j}, \bm{u'_j}), \label{eq:bilinear} \\
\notag &~~\text{where}~\bilinear(\bm{x}, \bm{y}) \triangleq \bm{x}^T \bm{W} \bm{y} + b.
\end{align}
$s_j$ and $s'_j$ are scores for positive and negative entities, respectively. 
The two BiLinear layers used in Eq.(\ref{eq:bilinear}) are parameter-tied.
We then use a margin-based hinge loss to train the MLM with the formulated pairwise ranking task, i.e., 
\begin{align}
\mathcal{L}_{R} = \dfrac{1}{m}\sum_{j=1}^{m} \max(\lambda - s_j + s'_j, 0), \label{eq:loss_ranking}
\end{align}
where the margin $\lambda$ is a hyperparameter.

The proposed distractor-suppressed ranking task has several nice properties. 
First, only a light-weight BiLinear layer is used to measure the score. 
Second, training to distinguish positive from negative samples may make the model more effective. 
Intuitively, two neighboring entities in graph are often assigned with similar distributed representations, but express differently in subtle context; this task helps discriminate them.
Finally, in contrast to the work of \citet{ye2019align}, ours is fully compatible with the entity-level MLM training task.


The final loss function for our model is defined as a combination of the entity-level MLM loss $\mathcal{L_{M}}$, and the distractor-suppressed ranking loss $\mathcal{L}_{R}$, with the latter weighted by a hyperparameter $\gamma$:
\begin{align}
\mathcal{L} = \mathcal{L}_{M} + \gamma \mathcal{L}_{R}. \label{eq:loss_all}
\end{align}


\subsection{Comparison to Prior Entity-Level MLMs}\label{sec:span_comparison}

Our work differs from prior entity-level MLMs, including SpanBERT \citep{pt/spanbert} and Baidu-ERNIE \citep{uc/baidu2019ernie,uc/baidu2019ernie2} in several ways. While the motivation of previous work is to move beyond token to another text unit, our method looks for ways to introduce structured knowledge from KGs into language models. As such, named entities in prior works are recognized via NLP toolkits. The entities are simply masked in random and relational knowledge unlikely exists among them.  
In contrast, entities in GLM are linked to a support knowledge graph, and masking has taken into consideration how an entity interacts with its neighbors in the KG. Similarly for modeling objective, previously proposed objectives such as span boundary objective  \citep{pt/spanbert}, aim at learning text semantics as in traditional MLM objectives. By exploiting relational knowledge among the recognized entities, 
we end up with a ranking task that is specially designed for the proposed entity-level MLM to directly acquire structured information.

\section{Experiments}
\subsection{Training Setup} \label{sec:exp_train_setup}

In this work we focus on non-canonicalized commonsense KGs, specifically ConceptNet, although the proposed approach is also applicable to ontological KGs such as  Freebase. 
For training efficiency we use two relatively small free-form corpora. One is the Open Mind Common Sense (OMCS) raw corpus\footnote{\url{https://github.com/commonsense}} consisting of 800K short sentences. 
The other is the ARC corpus~\citep{clark2018arc} containing 14M unordered, science-related sentences. 
Both corpora are parsed to have their entities linked to ConceptNet by using an inverted index built with fuzzy matching (Appx. A). 

For downstream tasks, CommonsenseQA~\citep{db/cqa} and SocialIQA~\citep{sap2019socialiqa} are used to evaluate GLM's performance on natural question answering (QA) task.
We also experiment with three knowledge base completion (KBC) tasks: WN18RR \citep{gnn/ConvE}, WN11 \citep{gnn/TransE} and commonsense knowledge base completion \citep{db/ckbc}, to assess whether the proposed approach can benefit graph-related tasks. 
We do not use other benchmarks like FB15k-237 \citep{gnn/ConvE} because they are derived from ontological KGs. 
And we do not use WN18 \citep{gnn/NTN} as it suffers from the ``informative value'' problem~\citep{gnn/ConvE}. 
Statistics of these benchmarks are summarized in Table~\ref{tab:dataset_stat}. 
\begin{table}[t] 
	\centering
	\begin{tabular}{lccccc}
		\toprule
		\bf{Dataset}& \bf{\# Entity}	& \bf{\# Rel}& \bf{\# Train}& \bf{\# Dev} & \bf{\# Test} \\
		\midrule
		{\footnotesize CommonsenseQA} & / & / & 9,741 & 1,221 & 1,140\\
		SocialIQA & / & / & 33,410 & 1,954 & 2,224\\
		\midrule
		WN18RR& 40,943 & 11 & 86,835 & 3,034 & 3,134\\
		WN11 & 38,696 & 11 & 112,581 & 2,609  & 10,544\\ 
		CKBC & 78,334  & 34 & 100,000 & {1,200/1,200} & 2,400 \\
		\bottomrule
	\end{tabular}
	\caption{\small Summary statistics of five benchmarks. The first two are multiple-choice question answering tasks. The rest include one link prediction and two triple classification tasks.}
	\label{tab:dataset_stat}
\end{table}
It is worth mentioning that, although WordNet~\citep{miller1998wordnet} is included in ConceptNet, the triples in ConceptNet are never used during GLM training but only raw texts from OMCS (which is a standalone source of ConceptNet and independent of WordNet), so the relation labels in WordNet are never seen by the GLM.  

For efficiency considerations, we initialized GLM with either BERT or RoBERTa rather than training from scratch. We choose to match the corresponding baseline model (whether it uses BERT or RoBERTa) in each downstream task for fair comparison. In practice, we can initialize GLM with any state-of-the-art pre-trained bidirectional language model. 

More training setups are detailed in Appendix A. 

\subsection{Question Answering Task Evaluation}\label{sec:qa_eval}

\paragraph{CommonsenseQA.} 
Table~\ref{tb:results_cqa} reports test results from the leaderboard\footnote{\url{www.tau-nlp.org/csqa-leaderboard}, and we thank Alon Talmor and Jonathan Herzig for test evaluations.} and from our approach. A brief introduction to each approach without reference can be found on the leaderboard. Compared to the previous best model RoBERTa+KE which is also trained with an extra in-domain corpus (i.e., OMCS) and uses retrieval during finetuning, our approach achieves 0.8\% absolute improvement to deliver a new state-of-the-art result. In addition, GLM based on RoBERTa-large outperforms its respective baseline RoBERTa-large by 2.0\%.

\begin{table}[t] 
	\centering
	\begin{tabular}{lcc} \toprule
		\textbf{Method} & \textbf{Dev} &\textbf{Test} \\ 
		\midrule 
		\midrule
		\multicolumn{3}{l}{	\textit{Models from pre-trained language model finetuning}}\\
		\midrule
		BERT-large \citep{pt/bert} & -& 56.7\\
		XLNet-large \citep{pt/xlnet} & -& 62.9\\
		RoBERTa-large \citep{pt/roberta} & 78.5& 72.1\\
		\midrule
		\midrule
		\multicolumn{3}{l}{	\textit{Models w/ IR$^{*}$ or extra supervisions during finetuning}}\\  
		\midrule
		CoS-E \citep{uc/rajani2019explain}& -& 58.2\\
		AristoBERTv7 (BERT-large)& -& 64.6\\
		DREAM (XLNet-large) & -& 66.9\\
		RoBERTa + IR& 78.9& 72.1\\
		RoBERTa + KE & 78.7& \underline{73.3}\\
		\midrule
		\midrule
		\multicolumn{3}{l}{	\textit{Models w/ further self-supervision tasks}}\\
		\midrule
		BERT+AMS \citep{ye2019align}& -& 62.2\\
		BERT+OMCS & -& 62.5\\
		RoBERTa+CSPT& 76.2& 69.6\\
		FreeLB-RoBERTa \citep{zhu2019freelb} & 78.8& 72.2\\
		\midrule 
		GLM (RoBERTa)$\dagger$ & \textbf{79.8} &\textbf{ 74.1}  \\
		\bottomrule 
	\end{tabular} 
	\caption{\small Results on CommonsenseQA for single models.``-'' denotes unavailable result, and underlined score is the previous best. 
		$^{*}$IR stands for information retrieval. 
		$\dagger$GLM is initialized with RoBERTa-large and falls into the last group.
	}
	\label{tb:results_cqa}
\end{table}

Note that approaches that use information retrieval (e.g., RoBERTa+KE) must retrieve from Wikipedia during finetuning and inference, which increases the computational overhead significantly. In contrast, approaches based on additional self-supervised pre-training are more efficient, but often achieve sub-optimal performance since they lack explicitly retrieved, targeted context. The proposed GLM falls into the latter high-efficiency group while still outperforms IR-based approaches. 

Some prior works (e.g., RoBERTa+CSPT) find that directly finetuning on triples from a KG can hurt performance. This evidence empirically supports our hypothesis that finetuning on triples can over-adapt a model to graph-based tasks and limit their generalization to other tasks.

Our approach is not directly comparable to \citet{uc/lv2019graph} and \citet{uc/lin2019kagnet}, the first of which achieves 75.3\% accuracy. This is because during finetuning and inference, those methods explicitly find a path from question to answer concept in ConceptNet. 
This helps filter human-generated distractor answers since they never appear in ConceptNet. In addition, our method never uses ConceptNet during finetuning and only observes a small subgraph of ConceptNet (about $30\%\!\sim\! 40\%$ linked concepts without relations) during pre-training.

\begin{table}[t] 
	\centering
	\begin{tabular}{lcc} \toprule
		\textbf{Method} & \textbf{Dev}& \textbf{Test}  \\ \midrule
		GPT \citep{pt/gpt} &  63.3& 63.0\\
		BERT-base \citep{pt/bert} & 63.3& 63.1\\
		BERT-large \citep{pt/bert} & 66.0& 64.5\\
		RoBERTa-large \citep{pt/roberta}  & 78.2& 77.1\\
		McQueen (RoBERTa) (Anonymous) & 79.5& {78.0}\\
		GB-KSI (Anonymous) & 77.5& \underline{78.1}\\
		\midrule
		GLM (RoBERTa)  &\textbf{79.6}&\textbf{78.6} \\ 
		\bottomrule 
	\end{tabular} 
	\caption{\small Results on SocialIQA. The results for comparative methods are copied from \citet{sap2019socialiqa} or leaderboard.}
	\label{tb:results_sqa}
\end{table}

\paragraph{SocialIQA.} The dataset\footnote{\url{https://tinyurl.com/socialiqa}} is built upon the ATOMIC knowledge graph \citep{db/atomic} and focuses on reasoning about people's actions and their social implications. 
It thus serves as an out-of-domain evaluation task for GLM trained using ConceptNet. 
Similar to CommonsenseQA, this task is formulated as a multiple-choice QA problem. 
The evaluation results listed in Table~\ref{tb:results_sqa} demonstrate that our approach also achieves state-of-the-art performance on this out-of-domain dataset. 

\begin{table}[t] 
	\centering
	\begin{tabular}{l|ccccc}
		\hline
		\multirow{2}*{\textbf{Method}} &\multicolumn{5}{c}{\textbf{WN18RR}} \\ 
		\cline{2-6}
		&MR &MRR &H@1 &H@3 &H@10  \\ 
		\hline
		TransE \citep{gnn/TransE}$\dagger$  & 2300& .243& .043& .441& .532 \\ 
		DistMult \citep{gnn/DistMult}$\dagger$ & 7000& .444& .412& .470& .504 \\ 
		ComplEx \citep{gnn/ComplEx}$\dagger$ & 7882& .449& .409& .469& .530 \\
		R-GCN \citep{gnn/RGCN}$\dagger$ & 6700& .123& .080& .137&  .207\\
		ConvE \citep{gnn/ConvE}$\dagger$ & 4464& .456& .419& .470& .531 \\ 
		ConvKB \citep{gnn/ConvKB}$\dagger$ & 1295& .265& .058& .445&  .558\\ 
		KBGAN \citep{cai2018kbgan} & -& .215& -& -&  .469\\ 
		RotatE \citep{gnn/RotatE}&3340& \underline{.476}& \underline{.428}& \underline{.492}& .571 \\ 
		\citet{uc/nathani2019learning}& 1940 & .440& .361& .483& \underline{.581} \\ 
		KG-BERT \citep{yao2019kgbert} & \underline{97} & -& -& -& .524 \\ 
		\hline
		KG-BERT (ours)$^{*}$ & 89& .474 & .356& .543& .678 \\ 
		GLM (BERT-base)  & \textbf{83}& \textbf{.557} & \textbf{.505}& \textbf{.552}& \textbf{.681} \\ 
		\hline
	\end{tabular}
	\caption{\small Link prediction results on the WN18RR benchmark. 
		$^{*}$We re-implemented data processing and ranking basis (detailed in Appx.~A). 
		$\dagger$Numbers are copied from \citet{uc/nathani2019learning}. 
	}
	\label{tb:results_lp}
\end{table}

\subsection{Graph-Related Task Evaluation}\label{sec:graph_eval}
For this set of tasks we follow KG-BERT \citep{yao2019kgbert} that finetunes BERT encoder over a concatenation of a triple's head, relation, and tail, followed by an MLP to compute a confidence score whether the triple is reasonable. Since BERT-base model is used in KG-BERT, for fair comparison we train a GLM from BERT-base, denoted as ``GLM (BERT-base)'' and finetune the GLM on KBC datasets. 

\paragraph{WordNet Knowledge Base Completion.}

Table~\ref{tb:results_lp} lists test results for the WN18RR link prediction task. GLM outperforms KG-BERT by a large margin and sets a new state-of-the-art result. GLM is superior to translation-based graph embedding models (e.g., RotatE and TansE), graph convolutional networks (e.g., R-GCN), and convolution-based methods (e.g., ConvKB and ConvE). 

\begin{table}[t] 
	\centering
	\begin{tabular}{lcc} \toprule
		\textbf{Method} & \textbf{Dev}& \textbf{Test}  \\ \midrule
		TransE~\citep{gnn/TransE} & -&75.9 \\   
		NTN~\citep{gnn/NTN}& -&86.2 \\
		DistMult~\citep{gnn/DistMult} & -&87.1\\
		DistMult-HRS~\citep{gnn/DistMult} & -&88.9 \\
		DOLORES~\citep{gnn/DOLORES} & -&87.5   \\
		ConvKB~\citep{gnn/ConvKB} & -&87.6 \\
		AATE~\citep{gnn/AATE} & -&88.0   \\    
		KG-BERT (BERT-base) \citep{yao2019kgbert} &-& {93.5} \\ \midrule
		GLM (BERT-base)  &\textbf{94.8}&\textbf{94.0} \\ 
		\bottomrule 
	\end{tabular} 
	\caption{\small Test accuracy on WN11 triple classification task.}
	\label{tb:results_wn11}
\end{table}

\begin{table}[t]  
	\centering
	\begin{tabular}{lcc} \toprule
		\textbf{Method} & \textbf{Dev2} &\textbf{Test} \\ \midrule
		Bilinear AVG \citep{db/ckbc} & 90.3 &91.7 \\ 
		DNN AVG \citep{db/ckbc} & 91.3 &92.0 \\ 
		Bilinear AVG + Data \citep{db/ckbc} & 91.8 &92.5 \\ 
		\midrule
		KG-BERT (BERT-base)& 92.9  & 93.2 \\
		KG-BERT + Data (BERT-base)& 92.3  & 92.4 \\
		\midrule
		GLM (BERT-base)  & 93.0 & 93.5 \\
		GLM (RoBERTa-large)  & \textbf{94.7} &\textbf{94.6} \\ 
		\bottomrule 
	\end{tabular} 
	\caption{\small Test accuracy on CKBC triple classification task.. } 
	\label{tb:results_ckbc}
\end{table}

Test results for the WN11 triple classification task are listed in Table~\ref{tb:results_wn11}. Consistent with the results on WN18RR, our approach outperforms translation-based graph embedding models and convolution-based methods, and improves state-of-the-art accuracy by 0.5\%. 

\paragraph{Commonsense Knowledge Base Completion (CKBC).} Finally, we evaluate our approach on the CKBC task, which should directly benefit from commonsense knowledge. Since CKBC is derived from the OMCS corpus, for fair comparison, we provide baseline model with equivalent training data (``+ Data'' in Table~\ref{tb:results_ckbc}). In addition, we remove raw sentences belonging to CKBC's test set from our GLM training corpora to avoid data leakage.

Results in Table~\ref{tb:results_ckbc} show our approach outperforms KG-BERT baseline even when the latter is equipped with equivalent data (increasing training triples from 100K to 600K). There are two possible reasons why performance actually drops with more data: 
1) the training triples are sorted w.r.t annotated confidence, so additional triples have a lower quality and may introduce noise; 
2) more negative sampling must be done with more training triples, which introduces more false negative examples (from 1.25\% to 2.42\%).

\begin{table}[t] 
	\centering
	\begin{tabular}{lc} \toprule
		\textbf{Method} & \textbf{F1 Score}\\ 
		\midrule
		BERT-large \citep{uc/feldman2019commonsense} & 78.8\\
		\midrule
		BERT-large (our implementation)& 77.1\\
		GLM (BERT-large) & \textbf{83.4}\\
		\bottomrule 
	\end{tabular} 
	\caption{\small Zero-shot evaluation by following \citet{uc/feldman2019commonsense}. }
	\label{tb:zero_shot}
\end{table}
\begin{table}[t]  
	\centering
	\begin{tabular}{lc} \toprule
		\textbf{Method} & \textbf{Dev Accu}\\ 
		\midrule
		BERT-large (w/o 2nd step pre-training) & 65.5\\
		\midrule
		GLM (BERT-large) & 69.0 \\
		$\Diamond$ w/o KG-guided Masking & 68.6 \\ 
		$\Diamond$ w/o Ranking & 68.1 \\ 
		$\Diamond$ w/o Ranking \& Entity Mask (= BERT-large) &  67.7 \\ 
		$\Diamond$ Ranking$\rightarrow$SBO &  66.8 \\ 
		$\Diamond$ Ranking$\rightarrow$SBO, Entity Mask$\rightarrow$Span &  68.2 \\ 	
		\bottomrule 
	\end{tabular} 
	\caption{\small Ablation study on CommonsenseQA dev set. Note, ``SBO'' denotes span boundary objective \citep{pt/spanbert} and ``$\rightarrow$'' stands for module replacement. }
	\label{tb:results_ablation}
\end{table}

\subsection{Zero-Shot Evaluation on CKBC}
To explore whether GLM can indeed learn the structured information from raw text, we conduct a zero-shot evaluation on CKBC by following \citet{uc/feldman2019commonsense}. For fair comparison, we re-train the GLM (BERT-large) on a new corpus in which all raw texts containing the CKBC testing pairs are discarded. We re-implement coherency ranking and estimate PMI \citep{uc/feldman2019commonsense}, and the results are shown in Table \ref{tb:zero_shot}. Our augmented pre-trained language model significantly outperforms its baseline, which demonstrates our model's capability of retaining structured information in a language model.

\subsection{Ablation Study}

To systematically evaluate the effectiveness of each component in the proposed approach, 
we conduct an ablation study in Table~\ref{tb:results_ablation} through pre-training language models with different setups and then finetuning on CommonsenseQA task. 

When KG-guided entity masking introduced in \S\ref{sec:node_lm} is replaced with random entity masking during GLM pretraining, 0.4\% accuracy drop is observed when the model is subsequently evaluated on CommonsenseQA dev set. If we set $\gamma$ in Eq.(\ref{eq:loss_all}) to zero when pre-training GLM, this leads to 0.9\% accuracy drop. When both entity-level masking and distractor-suppressed ranking are removed, the setting becomes equivalent to performing continual pre-training of BERT-large on our corpora. 
This helps us separate the contribution from extra in-domain data from the proposed approach. Compared with BERT-large baseline, we observe a 2.2\% (65.5\% to 67.7\%) accuracy improvement contributed by the corpora. Thus GLM yields an extra 1.3\% (67.7\% to 69.0\%) improvement by better data exploitation. 

When we replace distractor-suppressed ranking with span boundary objective~\citep{pt/spanbert}, a significant performance decrease (-2.2\%) is observed. Further replacing our entity-level masking with random span masking, however, only loses 0.8\% in accuracy. It is worth noticing that the latter setup is equivalent to continual pre-training with SpanBERT \citep{pt/spanbert} on our corpora. In line with SpanBERT's conclusion that the performance of linguistic masking is not consistent, our KG-guided entity-level MLM (i.e., GLM w/o the ranking task) is worse than random span with SBO (68.1\% vs. 68.2\%). 
This suggests that objective to mask and objective to learn need to be paired, and linguistic masking can be useful if equipped with appropriate learning objective (e.g., our distractor-suppressed ranking task) during pre-training.

\subsection{Analysis of Training \& Masking Schemes}

\begin{figure}[t]\small
	\centering
	\includegraphics[width=0.7\textwidth]{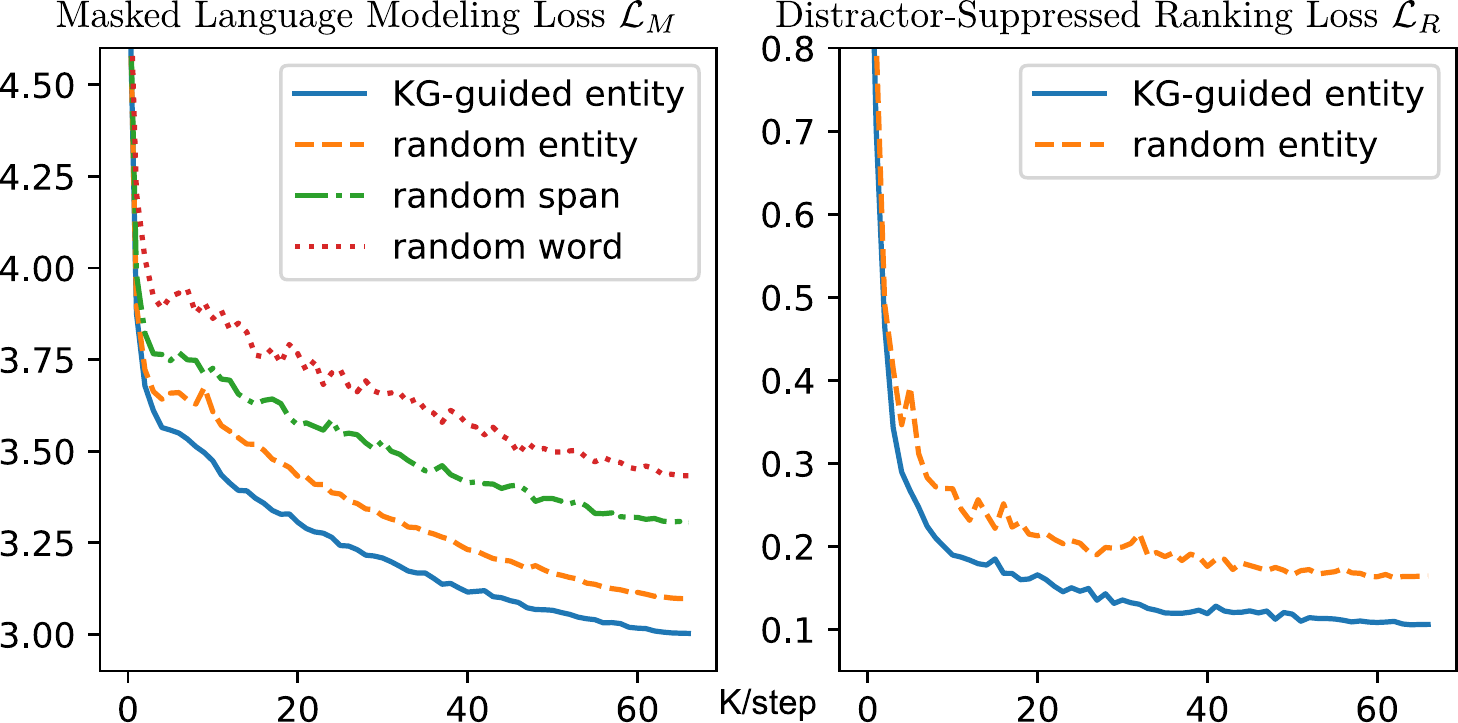}
	\caption{\small Dev loss during training phase for different training or masking schemes. Note only entity-level masking has $\mathcal{L}_{R}$. }
	\label{fig:training_loss} 
	\centering
\end{figure}

Given the same dev set of texts with the same masked tokens, Figure~\ref{fig:training_loss} compares different learning and masking schemes by plotting $\mathcal{L}_{M}$ (left) and $\mathcal{L}_{R}$ (right) defined in Eq.(\ref{eq:loss_all}) w.r.t training steps. It is observed that our KG-guided entity masking is more efficient than three other masking schemes, including random entity masking \citep{uc/baidu2019ernie}, random span masking \citep{pt/spanbert}, and random whole-word masking \citep{pt/bert}. 

Table~\ref{tb:case_study_mask} lists a few example sentences with masked tokens highlighted according to the corresponding masking scheme. Both KG-guided and random entity masking can mask informative chunks and long-term dependency needs to be modeled in order to infer the masked tokens. In contrast, random span or word masking is likely to mask tokens that can be easily inferred from local context -- a much simpler task. Furthermore, our KG-guided entity masking tends to select more informative phrases compared to random entity masking. 



\section{Related Work}  
Our work is related to Baidu-ERNIE \citep{uc/baidu2019ernie} and SpanBERT \citep{pt/spanbert}, which both extend token level masking to the span level. For example, Baidu-ERNIE does so to improve the model's knowledge learning, using uniformly random masking for phrases and entities. A detailed comparison between our model and span-level pre-trained language models can be found in \S\ref{sec:span_comparison}.
As briefly summarized in \S\ref{sec:introduction}, existing methods for integrating knowledge into pre-trained LMs can be coarsely categorized into two classes. For example, \citet{uc/peters2019knowledge} retrieve entities' embeddings according to the similarity between a Transformer's hidden states and pre-trained graph embeddings, then treat the retrieved embeddings as extra inputs to the next layer. 
In contrast, \citet{uc/bosselut2019comet} directly finetune a pre-trained LM on partially-masked triples from a KG, aiming at commonsense KBC tasks.  
In addition our work is also related to using negative samples for effective learning  \citep{cai2018kbgan}. 

Our work also differs from the works combining knowledge graph with text information via joint embedding \cite{yamada2016joint}. 
They usually use the texts containing co-occurrence of entities to enrich the graph embeddings, which are specially designed for graph-related tasks.  
For example, \cite{wang2014knowledge} embed entities from KG and the entities' text contents in the same latent space, however, regardless of textual co-occurrences and their textual relations in natural language corpus. 
Further taking into account the sharing of sub-structure in the textual relations in a large-scale corpus, \cite{toutanova2015representing} apply a CNN to the lexicalized dependency paths of the textual relation, for an augmented relation representation. The representation can be fed into any previous graph embedding approach for enhanced performance on KBC. 
We share similar inspirations when utilizing the texts containing entity co-occurrences and embedding entities' text contents into latent space. 
But beyond the shallow joint embeddings, our work takes advantage of pre-trained MLMs and equips them with structured knowledge via two self-supervised objectives built upon raw text. Hence it can produce generic text representations to benefit various downstream tasks.

\begin{table}[t]  \small
	\centering
	\begin{tabular}{p{0.4cm}p{6.3cm}} \toprule
		\textbf{No.}& \textbf{Masked Text with different masking methods}\\ 
		\midrule
		1& something you \uwave{need to \textit{do}} \textit{before} \textit{\textbf{you}} \uline{get up \textbf{early}} is \textbf{set} an alarm $\leftarrow$ (leave the office) \\
		2& you would talk \textbf{with} someone \uwave{far} \textit{\textbf{away}} \textit{because} \textit{you} \uwave{want }\uline{\uwave{keep in} touch } $\leftarrow$ (meet strangers) \\
		3& \uwave{if you want} \textbf{to} \uline{drill \textit{a} \textit{\textbf{hole}}} \textit{\textbf{then}} you \textbf{should} carefully plan \textbf{where} you will drill it $\leftarrow$ (cut of beef) \\
		\bottomrule 
	\end{tabular} 
	\caption{\small Case study for different masking schemes. Note that 1) \uline{underline}: KG-guided entity masking;  2) \uwave{underline wave}: random entity masking; 3) \textit{italic}: random span masking; and 4) \textbf{bold}: random whole-word masking. Text in parenthesis is a negative sample for KG-guided entity masking. }
	\label{tb:case_study_mask}
\end{table}

\section{Conclusion and Discussion} 

In this work, we aim at equipping pre-trained LMs with structured knowledge through novel self-supervised tasks. 
Building upon entity-level MLMs, we propose an entity masking scheme under guidance from a KG. This method masks informative entity mentions and facilitates learning structured knowledge underlying free-form text. 
In addition, we propose a distractor-suppressed ranking objective to utilize negative samples from KG as distractors for effective model training. 
Experiments show that finetuning our KG-guided pre-trained MLMs yields improved performance on relevant downstream tasks. 
In the future, we will use a combination of commonsense and ontological KGs, and large-scale corpora (e.g., Wikipedia or Common Crawl) to pre-train an MLM from scratch, which we expect to benefit a wide range of tasks. 

\bibliographystyle{plainnat}
\bibliography{ref}

\appendix

\section{Implementation Details}\label{sec:details}

Pre-trained Transformer-based language models implemented by Huggingface\footnote{\url{https://github.com/huggingface/transformers}} are adapted for our models. 

\paragraph{Training Hyperparameters.} 
For continual pre-training, we do not tune hyperparameters due to the computational cost. Only a limited set of hyperparameters are tried according to empirical intuitions. Currently $R_{hop/min/max}$ in Eq.(\ref{eq:prob_heuristic}) are set to 3/1.0/2.0 respectively, and $R_{thresh}$ aims to filter out entities with top 5\% document frequency, thus varies with corpora. We set $\lambda$ in Eq.(\ref{eq:loss_ranking}) to be 1.0 and $\gamma$ in Eq.(\ref{eq:loss_all}) to be 0.2. The continual pre-training runs for 5 epochs, with a batch size of 128, a learning rate of 3e-5/1e-5 (BERT/RoBERTa), learning warmup proportion of 10\%/5\% (BERT/RoBERTa) and weight decay of 0.01.
For both BERT and RoBERTa, max sequence length is set to be 80 and 20 for sentence-level encoding and entity embedding respectively. The masking proportion is lifted from 15\% to 20\% without tuning compared with BERT and RoBERTa. An intuition is that our model is initialized with well-trained language models (e.g., BERT), a slightly larger masking proportion could hold the entity with longer text span, and make the learning more efficient.

During finetuning on downstream tasks, we conduct grid search for hyperparameters, including batch size, number of epochs/steps, learning rate, which are summarized in Table~\ref{tb:hparams_search}. 

\begin{table*}[htbp]\small
	\centering
	\setlength{\tabcolsep}{.05pt}
	\begin{tabular}{@{}ll|ccccc@{}}
		\toprule
		\textbf{Hparam}  & \textbf{GLM Model} & \textbf{CQA} & \textbf{SocialIQA} & \textbf{WN18RR} &  \textbf{WN11} & \textbf{CKBC} \\ \midrule 
		\multirow{2}{*}{Batch Size}     & BERT      &     \{8, 12, 16\}&  /&     \{32\} &        \{32, 48, 64\}&     \{24 32\} \\ 
		& RoBERTa   &     \{8, 12, 16\}&   \{12, 16\}&      /&            /&      \{24 32\}\\ \midrule 
		\multirow{2}{*}{\# Steps/Epochs} & BERT      &     \{2800, 3400, 4000\}&      /&   \{5\}&         \{3, 5\} &      \{6\}\\ 
		& RoBERTa   &     \{2800, 3600\}&    \{3,4\}&     /&       /&       \{6\}\\ \midrule 
		\multirow{2}{*}{Learning Rate}  & BERT      &    \{3e-5, 5e-5, 7e-5\} &     /&    \{5e-5\}&          \{3e-5, 5e-5, 7e-5\}&      \{3e-5, 5e-5, 7e-5\}\\ 
		&  RoBERTa  &    \{1e-5, 8e-6\} &    \{1e-5, 8e-6\}&     /&          /&      \{1e-5, 2e-5\}\\ \bottomrule 
	\end{tabular}
	\caption{Candidate values for hyperparameter search. Note ``\# steps'' is shown for CQA and ``\# epochs'' for the others.
	}
	\label{tb:hparams_search}
\end{table*}

\paragraph{Details about KG-Guided Entity Masking.} 
In addition to finding informative and non-trivial masks, the KG-guided entity masking method can be used to filter the corpus, since if a piece of text (e.g., sentence or passage) contains only trivial and undeducible entities, the text may hardly contribute to model learning. With this strategy, most ($\sim 90\%$) sentences in ARC corpus were filtered out, which led to a very efficient training for our models. Training our model based on BERT-large or RoBERTa-large only costs about 1 day on single V100 GPU with mixed float precision, in total about 70K steps. 

\paragraph{Fair Comparison on WN18RR.} There are two aspects to prevent data leakage from ConceptNet to WN18RR. First, ConceptNet is derived from OMCS, WordNet, Wiktionary, etc., and each source is independent of others. Our method only uses OMCS raw text as the training corpus and has no access to WordNet. On the other hand, even though entire ConceptNet is used to guide entity masking/sampling, unlike traditional methods taking KG's triples as training examples, our method doesn't use the relation labels but only the entities/concepts. Since WordNet's entities/concepts are also provided in the training set of WN18RR/WN11, there is no risk of leakage when testing. 

\paragraph{KG-BERT Re-Implementation.} 
In this work, we re-implemented KG-BERT \citep{yao2019kgbert} for improving the efficiency of test inference and negative sampling, as the original implementation is less optimized. The original version requires about 3 days to make a full linking prediction inference on the test set. Our re-implemented version reduces the time to 20 hours (i.e., 3$\times$ acceleration) using the same single GPU. We also improved negative sampling with $\sim 3 \times$ speed than the original version. The modifications can be summarized as 1) the ranking basis is changed from logits to probabilities; 2) mixed float is employed to replace the float; and 3) negative sampling is re-implemented for efficiency. 
In addition, we found that, in any evaluation script \citep{yao2019kgbert,gnn/RotatE}, the model can successfully rank the candidates as long as assigning the positive triple with one of the largest scores. 
The following can explain why the re-implemented KG-BERT can achieve much superior performance. First, the mixed float normalized probabilities, rather than logits, are used as ranking basis, which only ranges from 0 to 1. And second, such encoder and classifier paradigm (e.g., KG-BERT) for link prediction usually suffers from polysemy or ambiguity of entities, which leads to over-confident false positive prediction. Hence, our re-implementation significantly improves the Hits@1 for KG-BERT, which however is still a reasonable setting to verify and compare how much structural knowledge is retained in the pre-trained LMs. For a fair comparison with previous works, we list the results with ranking basis of logits in Table \ref{tb:results_lp_new}.
\begin{table}[t] 
	\centering
	\begin{tabular}{l|ccccc}
		\hline
		\multirow{2}*{\textbf{Method}} &\multicolumn{5}{c}{\textbf{WN18RR}} \\ 
		\cline{2-6}
		&MR &MRR &H@1 &H@3 &H@10  \\ 
		\hline
		KG-BERT (BERT-base) \citep{yao2019kgbert} & 97 & .216 & .041& .302& .524 \\ 
		GLM (BERT-base)  & \textbf{86}& \textbf{.273} & \textbf{.086}& \textbf{.344}& \textbf{.587} \\ 
		\hline
	\end{tabular}
	\caption{\small Link prediction results on the WN18RR benchmark with ranking basis of logits, instead of probabilities in the main paper. 
	}
	\label{tb:results_lp_new}
\end{table}

\paragraph{KBC Tasks Implementation.} 
For CKBC dataset, we find the performance is very poor when we directly finetune either BERT-base or our approach on the concatenated triples, i.e., ``\texttt{[CLS]} \textsc{Head} \texttt{[SEP]} \textsl{Rel} \texttt{[SEP]} \textsc{Tail} \texttt{[SEP]}''. Hence, we follow \citet{uc/feldman2019commonsense} to transform the triples to natural language sentences, and then use these sentences as input to finetune the pre-trained LMs. 
For WN18RR, we directly use the data processed by \citet{yao2019kgbert}, in which a description sentence is attached to each entity/phrase. 
For WN11, we follow KG-BERT to directly concatenate the triples and then use them to finetune the pre-trained LMs. 

\paragraph{Knowledge Graph and Entity Linking.} 
We aim at enhancing the pre-trained language models with commonsense structured knowledge, so we employ ConceptNet \citep{db/conceptnet} as the backend knowledge graph. Since ConceptNet is a multi-lingual knowledge graph, we first filtered out all the triples which include non-English items. In addition, we treated the KG as an undirected graph when identifying entity's mutual reachability. 
As for entity linking, there are many mature entity linking systems for ontological or factoid KGs, such as S-MART \citep{yang2015smart}, DBpeida Lookup, and DeepType \citep{raiman2018deeptype}. However, for commonsense KG whose content consists of non-canonicalized or free-form texts, there is no such a system to complete its entity linking. Therefore, we built an efficient inverted index out of lemma-based fuzzy matching as our entity linking system 
which is going to be open-sourcing.

\paragraph{Evaluation Metrics.} 
For multiple-choice question answering tasks and triple classification tasks, we use accuracy as the metric. For link prediction task, there are two kinds of metrics: the first is mean rank (MR) and mean reciprocal rank (MRR), and the second is H@N (namely Hits@N) that means the proportion of correct entities in top N after being sorted w.r.t predicted confidence. 
Note we only report results under the \textit{filtered} setting \citep{gnn/TransE} which removes all corrupted triples appearing in training, dev and test set.

\section{More Experiments}

\paragraph{PhysicalIQA.}
In addition to CommonsenseQA and SocialIQA shown in the main paper, a dataset named PhysicalIQA\footnote{\url{https://leaderboard.allenai.org/physicaliqa/submissions/public}} is also used to evaluate our method. 
It is also regarded as an out-of-domain dataset compared with our training corpora. 
However, our implemented code base cannot reproduce the state-of-the-art results that are achieved by RoBERTa-large finetuning, possibly due to different pre-processing and feeding strategies for pre-trained LMs, e.g., special token, concatenation scheme, representation gathering method  \citep{mitra2019exploring}. 
Hence, we only report re-implemented results with the same network structure, same data-preprocessing method, same random seed and same hyperparameter grid search, for fair comparison. 
The accuracy on dev set is 78.7\% and 80.2\% for RoBERTa-large baseline and GLM (RoBERTa) respectively, which further demonstrates the effectiveness of our approach on out-of-domain datasets. 

\paragraph{HellaSWAG.}
We also try to apply our approach to HellaSWAG\footnote{\url{https://leaderboard.allenai.org/hellaswag}} \citep{zellers2019hellaswag}. 
It is a plausible inference task and requires reasoning over linguistic context and external knowledge. The task is to choose one plausible ending from four candidates. Same as PhysicalIQA, we only fairly report the accuracy on dev set for a fast comparison. With our implementation, finetuning RoBERTa-large on this dataset achieves 84.1\% dev accuracy which is much higher than the best dev accuracy (83.5\%) on leaderboard. However, finetuning our approach achieves 83.9\% accuracy, which is slightly worse than the baseline. 
We noticed that examples in HellaSWAG frequently have multiple consecutive sentences for inference, thus our model trained on single, unordered sentences may only achieve sub-optimal performance. On the other hand, this is another out-of-domain dataset which may not benefit from our training knowledge graph and corpora. 

\section{Error Analysis} \label{app:error_analysis}
Compared with original pre-trained LMs (e.g., BERT and RoBERTa), some limitations are found in our pre-trained models, which mainly fall into
\begin{itemize}
\item\textbf{Over masking:} Compared to random masking, our KG-guided masking scheme is more likely to mask all key parts of a sentence, which leaves little room for MLM task. 
\item\textbf{Short context:} Since our corpora consist of just single, unordered sentences, information that spans across multiple sentences is not encoded effectively. When downstream tasks rely heavily on consecutive sentences, finetuning our model yields inferior performance. This can be empirically verified by the performance drop on HellaSWAG dataset which involves long contexts. 
\item\textbf{Pipeline model:} Same as any other method aiming to integrate LMs with KG, a linking system is first applied to detect entities in a text, which inevitably suffers from graph sparsity and leads to error propagation. However, our method is less sensitive to such errors compared with the methods that links entity during both finetuning and inference.
\end{itemize}

\section{Related Work (extended)} \label{app:related_work_full} 

This work is in line with Baidu-ERNIE \citep{uc/baidu2019ernie} and SpanBERT \citep{pt/spanbert} which replace word-level mask \citep{pt/bert} with span-level one for knowledge information and long-term dependency. In particular, Baidu-ERNIE uses uniformly random masking for phrases and entities, whereas SpanBERT directly masks out token spans sampled under geometric distribution. 

The work of \citet{uc/petroni2019language} finds that, without finetuning, pre-trained LMs (e.g., BERT) contains relational knowledge competitive with traditional NLP methods with oracle knowledge. Nevertheless, how to integrate the oracle knowledge into the pre-trained LMs for further performance improvement remains an open question. 

As briefly summarized in the introduction of the main paper, existing methods can be coarsely categorized into two classes. For the first class, those methods retrieve a KG subgraph or/and pre-trained graph embeddings via entity linking during finetuning and inference. 
K-BERT \citep{uc/liu2019kbert} retrieves a path from KG as description for each detected entity in text, and inserts such description into input sequence to Transformer encoder with carefully designed attention mask and position embedding. 
KnowBert \citep{uc/peters2019knowledge} and THU-ERNIE \citep{uc/thu2019ernie} first retrieve the detected entities' embeddings from pre-trained graph embeddings \citep{gnn/TransE}, and then treat these retrieved embeddings as extra inputs for each layer of Transformer encoder. 
\citet{uc/lin2019kagnet} and \citet{uc/lv2019graph} aim to solve commonsense multiple-choice QA problem. They retrieve a graph path from entities detected in question to each answer entry, and then encode (e.g., via LSTM) these paths as heterogeneous representations for higher-level modules. 

The second class of methods use contextualized representations from pre-trained LMs to enrich graph embeddings and thus alleviate graph sparsity issues. 
COMET \citep{uc/bosselut2019comet} finetunes pre-trained LM on partially-masked triples from KG, which only aims at commonsense knowledge graph completion tasks. 
\citet{uc/malaviya2019exploiting} perform transfer learning from pre-trained language models to knowledge graphs for enhanced contextual representation of the knowledge. 
KG-BERT \citep{yao2019kgbert} directly concatenates the head, relation and tail of a triple, and finetunes pre-trained LMs on such data with binary classification objective, i.e., whether a triple is correct or not. 

How to generate and utilize negative samples is important for learning graph embeddings and structured knowledge \citep{gnn/RotatE,ye2019align}. For example, 
KBGAN \citep{cai2018kbgan} uses a knowledge graph embedding model as negative sample generator to assist the training of the desired model, which acts as the discriminator in GANs.
Rather than a standalone generator, self-adversarial sampling \citep{gnn/RotatE} generates negative samples according to the current entity or relation embeddings. 
BERT-AMS \citep{ye2019align} and our proposed ranking task share a similar motivation that the model is able to effectively learn the structured knowledge from negative samples, but they differ in the task designs. BERT-AMS builds a multiple-choice question answering task for utilizing negative samples, which imitates the developing procedure of CommonsenseQA \citep{db/cqa} and aims to improve performance on that particular dataset. In contrast, our approach is more general and formulates a ranking task along with the entity-level masked language modeling objective for pre-training knowledge-aware LMs.


\end{document}